\pdfoutput=1

\documentclass[11pt]{article}

\usepackage{acl}
\usepackage{times}
\usepackage{comment}
\usepackage{latexsym}
\usepackage{todonotes}
\usepackage{enumitem}
\usepackage[T1]{fontenc}

\usepackage[utf8]{inputenc}

\usepackage{microtype}

\usepackage{inconsolata}

\usepackage{algorithm}
\usepackage{multirow}
\usepackage{algpseudocode}
\usepackage{hyperref}
\usepackage{xcolor}
\usepackage{amssymb} 
\usepackage{tcolorbox} 
\usepackage{xcolor}
\usepackage{graphicx}
\usepackage{pifont}
\usepackage{amsmath}
\usepackage{booktabs}
\usepackage{tabularx}
\usepackage{tikz}
\usetikzlibrary{shadows}
\usepackage{tcolorbox}
\usepackage{varwidth} 
\usepackage{caption}
\usepackage{subcaption}
\usepackage{ragged2e}
\usepackage{newunicodechar}
\newunicodechar{✓}{\checkmark}
\newunicodechar{✗}{\ding{55}}

\title{Are Large Language Models Economically Viable for Industry Deployment?}

\author{
\textbf{Abdullah Mohammad}\textsuperscript{1},
\textbf{Sushant Kumar Ray}\textsuperscript{2},
\textbf{Pushkar Arora}\textsuperscript{3},
\textbf{Rafiq Ali}\textsuperscript{4}\\
\textbf{Ebad Shabbir}\textsuperscript{5},
\textbf{Gautam Siddharth Kashyap}\textsuperscript{6},
\textbf{Jiechao Gao}\textsuperscript{7}\thanks{Corresponding Author: jiechao@stanford.edu, usman.naseem@mq.edu.au},
\textbf{Usman Naseem}\textsuperscript{8}\footnotemark[1] \\
\textsuperscript{1, 3, 4, 5}DSEU-Okhla, New Delhi, India \\[-0.4ex]
\textsuperscript{2}University of Delhi, New Delhi, India \\[-0.4ex]
\textsuperscript{6, 8}Macquarie University, Sydney, Australia \\[-0.4ex]
\textsuperscript{7}Center for SDGC, Stanford University, California, USA \\
}

\usepackage{textcomp}
\begin{document}
\maketitle
\begin{abstract}
Generative AI—powered by Large Language Models (LLMs)—is increasingly deployed in industry across healthcare decision support, financial analytics, enterprise retrieval, and conversational automation, where reliability, efficiency, and cost control are critical. In such settings, models must satisfy strict constraints on energy, latency, and hardware utilization—not accuracy alone. Yet prevailing evaluation pipelines remain accuracy-centric, creating a \emph{Deployment--Evaluation Gap}—the absence of operational and economic criteria in model assessment. To address this gap, we present \textsc{Edge-Eval}\footnote{\url{https://github.com/Abdullah4152/EDGE-EVAL}}—a industry-oriented benchmarking framework that evaluates LLMs across their full lifecycle on legacy NVIDIA Tesla T4 GPUs. Benchmarking LLaMA and Qwen variants across three industrial tasks, we introduce five deployment metrics—\emph{Economic Break-Even} ($N_{break}$), \emph{Intelligence-Per-Watt} ($IPW$), \emph{System Density} ($\rho_{sys}$), \emph{Cold-Start Tax} ($C_{tax}$), and \emph{Quantization Fidelity} ($Q_{ret}$)—capturing profitability, energy efficiency, hardware scaling, serverless feasibility, and compression safety. Our results reveal a clear efficiency frontier—models in the \emph{$<2$B parameter class} dominate larger baselines across economic and ecological dimensions. \emph{LLaMA-3.2-1B (INT4)} achieves ROI break-even in $14$ requests (median), delivers $3\times$ higher energy-normalized intelligence than $7$B models, and exceeds $6{,}900$ tokens/s/GB under 4-bit quantization. We further uncover an efficiency anomaly—while QLoRA reduces memory footprint, it increases adaptation energy by up to $7\times$ for small models—challenging prevailing assumptions about quantization-aware training in edge deployment.
\end{abstract}

\begin{figure}[t]
\centering
\includegraphics[width=\columnwidth]{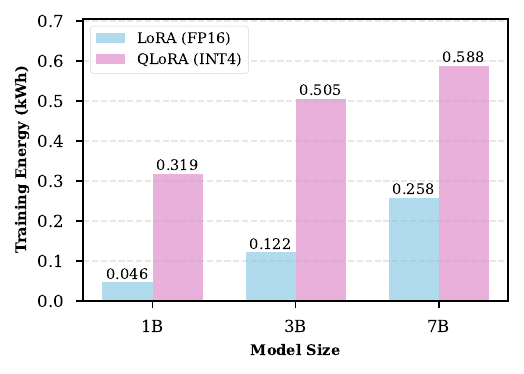}
\caption{Illustration of the \emph{Deployment--Evaluation Gap}--QLoRA reduces memory by $\sim60\%$ yet increases fine-tuning energy up to $7.2\times$ for small models, showing that memory efficiency does not equal energy efficiency.}
\label{fig:efficiency_gap}
\end{figure}

\section{Introduction}
\label{sec:intro}

Generative AI—powered by Large Language Models (LLMs) \cite{ciubotaru2025generative}—is rapidly transitioning from research prototypes to real-world industry deployment. Across healthcare decision support \cite{almadani2025systematic}, financial analytics \cite{al2023big}, enterprise retrieval \cite{hasan2022information}, and conversational automation \cite{manolescu2025interactive}, these models must operate under strict constraints on energy, latency, cost, and hardware availability. In such environments, practical viability depends not only on predictive accuracy, but also on economic sustainability. Despite this, prevailing evaluation pipelines remain dominated by accuracy-centric benchmarks \cite{siddiqui2025llms, joshi2025can, hendrycks2020measuring}. These benchmarks provide limited insight into operational trade-offs, creating what we term the \emph{Deployment--Evaluation Gap}—the absence of operational and economic criteria in model assessment. Figure~\ref{fig:efficiency_gap} illustrates a motivating example of this gap. While Quantized Low-Rank Adaptation (QLoRA) \cite{dettmers2023qlora} reduces memory usage by approximately $60\%$, it increases fine-tuning energy consumption by up to $7.2\times$ for small models compared to standard LoRA \cite{hu2022lora}. Memory efficiency, therefore, does not necessarily translate to energy efficiency. Such trade-offs remain invisible under accuracy-centric benchmarks, yet they critically impact real-world deployment decisions.

\begin{figure*}[t]
\centering
\includegraphics[width=0.85\linewidth]{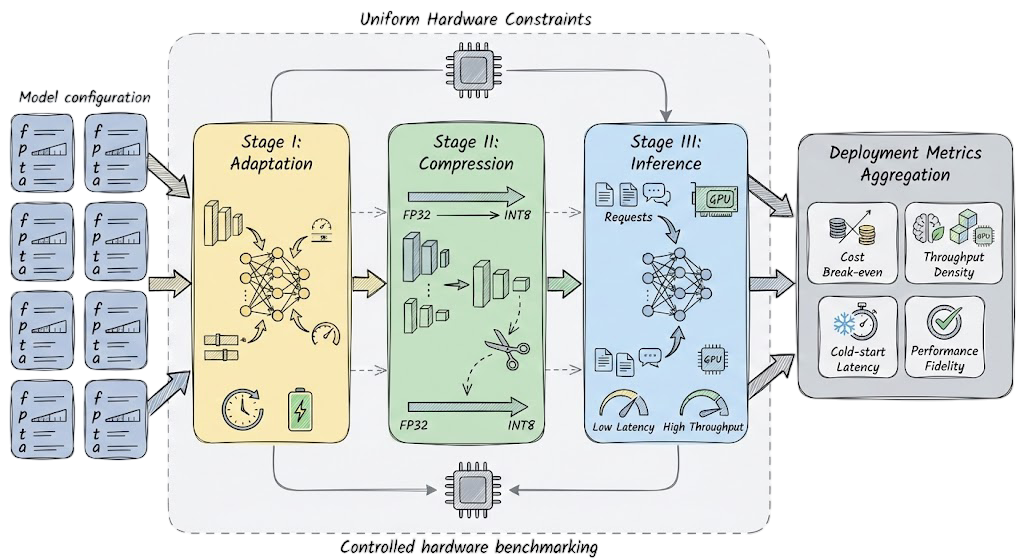}
\caption{Lifecycle benchmarking pipeline of \textsc{Edge-Eval}. For each configuration $(f,p,t,a)$, models pass through three stages—\emph{adaptation}, \emph{compression}, and \emph{inference}—under uniform hardware constraints. The recorded lifecycle variables are subsequently aggregated into the five deployment metrics defined in Section~\ref{evaluation}.}
\label{arc}
\end{figure*}

To address this gap, we introduce \textsc{Edge-Eval}—an industry-oriented benchmarking framework that evaluates language models across their full operational lifecycle. We conduct a empirical study of LLaMA \cite{grattafiori2024llama3herdmodels} and Qwen \cite{yang2024qwen25} variants across three industrial tasks—Summarization (long-context document compression under constrained inference budgets), Retrieval-Augmented Generation (RAG) \cite{lewis2020retrieval} (retrieval-grounded reasoning with external knowledge integration), and Conversational Agents (latency-sensitive, multi-turn instruction-following dialogue)—on widely deployed NVIDIA Tesla T4 hardware. Our methodology integrates LoRA, QLoRA, and vLLM-based inference \cite{kwon2023vllm}. Furthermore, \textsc{Edge-Eval} extends conventional benchmarking through five deployment metrics--(1) \emph{Economic Break-Even ($N_{break}$)}—the traffic volume required for local adaptation to undercut API costs; (2) \emph{Intelligence-Per-Watt ($IPW$)}—performance normalized by energy consumption \cite{schizas2022tinyml, patterson2022carbon}; (3) \emph{System Density ($\rho_{sys}$)}—throughput per gigabyte of VRAM; (4) \emph{Cold-Start Tax ($C_{tax}$)}—the energy penalty of model loading; and (5) \emph{Quantization Fidelity ($Q_{ret}$)}—reasoning retention under 4-bit compression. In summary, our work makes two primary contributions:
\begin{itemize}
 \vspace{-0.3cm}
    \item We introduce \textsc{Edge-Eval}, a benchmarking framework that augments accuracy-centric evaluation with five deployment metrics ($N_{break}$, $IPW$, $\rho_{sys}$, $C_{tax}$, $Q_{ret}$) for lifecycle assessment of LLMs on legacy hardware.
    \vspace{-0.8cm}
    \item Empirically, \textsc{Edge-Eval} on LLaMA and Qwen variants on Tesla T4 GPUs, we identify an efficiency frontier where $<2$B models outperform larger baselines, and reveal an energy anomaly—QLoRA increases adaptation energy by up to $7\times$ despite reducing memory footprint.
\end{itemize}

\section{Related Work}

\paragraph{Model Compression.}
Recent advances in model compression and Parameter-Efficient Fine-Tuning (PEFT) have enabled LLMs to operate on resource-constrained hardware. Post-Training Quantization (PTQ) models such as Generalized Post-Training Quantization (GPTQ) \cite{frantar2022gptq} and Activation Aware Quantization (AWQ) \cite{lin2024awq} reduce numerical precision while preserving accuracy, and Quantization-Aware Training (QAT) models such as QLoRA \cite{dettmers2023qlora} integrate low-bit quantization into adaptation. Similarly, LoRA \cite{hu2022lora} and related PEFT models freeze base weights and introduce low-rank adapters to reduce memory overhead \cite{lialin2023stack}. While these models demonstrate strong accuracy-centric. Yet, systematic analysis of lifecycle energy consumption, economic trade-offs, and infrastructure-level efficiency remains limited. In particular, memory reduction is often implicitly treated as a proxy for deployment efficiency—an assumption our empirical results challenge.

\paragraph{Model Evaluation.}
Green AI initiatives advocate reporting energy consumption alongside accuracy \cite{schizas2022tinyml, patterson2022carbon}, such as MLPerf Tiny \cite{banbury2021mlperf} evaluate inference on ultra-low-power devices. Other studies profile latency or throughput for quantized inference \cite{yao2022zeroquant}, yet typically omit economic viability, cold-start overhead, and hardware density considerations. Existing benchmarks therefore lack a unified framework that connects adaptation cost, inference energy, quantization fidelity, and return on investment. In contrast, \textsc{Edge-Eval} introduces a lifecycle-oriented evaluation paradigm through five deployment metrics ($N_{break}$, $IPW$, $\rho_{sys}$, $C_{tax}$, $Q_{ret}$), providing a systematic assessment of operational and economic criteria in model assessment.

\section{Methodology}
\label{sec:methodology}

\textsc{Edge-Eval} (see Figure \ref{arc}) benchmarks language models through a structured lifecycle pipeline that mirrors real-world deployment. Let $\mathcal{F}$ denote the set of model families (LLaMA, Qwen) described in Section~\ref{model}, $\mathcal{P}$ the set of parameter tiers (Micro $<2$B, Compact 3B, Standard 7B--8B), $\mathcal{T}$ the set of industrial tasks defined in Section~\ref{Datasets} (Summarization, RAG, Conversational Agents), and $\mathcal{A}$ the set of adaptation strategies (LoRA-FP16, LoRA-INT8, LoRA-INT4, QLoRA-INT4). For each configuration $(f,p,t,a) \in \mathcal{F} \times \mathcal{P} \times \mathcal{T} \times \mathcal{A}$, we execute a full deployment pipeline consisting of adaptation, compression (when applicable), and serving. This factorial design yields $|\mathcal{F}| \times |\mathcal{P}| \times |\mathcal{T}| \times |\mathcal{A}| = 72$ benchmarked variants. During the \emph{adaptation stage}, models are specialized on task-specific data using PEFT. In the \emph{compression stage}, weights are optionally quantized under controlled precision regimes. In the \emph{inference stage}, adapted models are deployed in a serving environment representative of low-batch industry conditions. For each stage, we record lifecycle variables including training energy $E_{\text{train}}$, inference energy per request $E_{\text{infer}}$, model loading overhead $E_{\text{load}}$, sustained throughput $T_{\text{put}}$, latency characteristics $(T_{\text{TTFT}}, T_{\text{ITL}})$, and GPU memory footprint $M_{\text{vram}}$. These measured quantities collectively characterize both one-time specialization cost and recurring operational behavior under uniform hardware constraints. The recorded variables are subsequently aggregated into five deployment metrics—$N_{break}$, $IPW$, $\rho_{sys}$, $C_{tax}$, and $Q_{ret}$—whose formal definitions and mathematical formulations are provided in Section~\ref{evaluation}.

\section{Experimental Setup}
\label{model}

All experiments are conducted on a dual-GPU node equipped with NVIDIA Tesla T4 accelerators (16GB VRAM each) \cite{nvidia2018nvidia, jia2019dissectingnvidiaturingt4}, reflecting widely deployed legacy industry hardware. We evaluate two open-weight model families—LLaMA \cite{grattafiori2024llama3herdmodels} (1B\footnote{\url{https://huggingface.co/meta-LLaMa/LLaMa-3.2-1B-Instruct}}, 3B\footnote{\url{https://huggingface.co/meta-LLaMa/LLaMa-3.2-3B-Instruct}}, 8B\footnote{\url{https://huggingface.co/meta-LLaMa/LLaMa-3.1-8B-Instruct}}; employing Grouped-Query Attention) and Qwen-2.5 \cite{yang2024qwen25} (1.5B\footnote{\url{https://huggingface.co/Qwen/Qwen2.5-1.5B-Instruct}}, 3B\footnote{\url{https://huggingface.co/Qwen/Qwen2.5-3B-Instruct}}, 7B\footnote{\url{https://huggingface.co/Qwen/Qwen2.5-7B-Instruct}}; dense transformer variants)—across three parameter tiers. Models are adapted using PEFT with rank $r=16$ and scaling factor $\alpha=32$, under four precision configurations--LoRA-FP16, LoRA-INT8, LoRA-INT4 (PTQ), and QLoRA-INT4 \cite{hu2022lora, dettmers2023qlora}. Inference is served via vLLM (v0.6.3) \cite{kwon2023vllm} with paged attention enabled, and evaluated under batch size 1 to simulate low-batch deployment conditions. Throughput (tokens/s), Time-To-First-Token (TTFT), and Inter-Token Latency (ITL) are measured over 100 independent requests per configuration, while GPU power draw is recorded using \texttt{pynvml} \cite{bauer2024greem} at 100\,ms intervals to enable fine-grained lifecycle energy profiling.

\begin{table*}[t]
\centering
\scriptsize
\resizebox{\textwidth}{!}{%
\begin{tabular}{@{}llccccc@{}}
\toprule
\textbf{Family} & \textbf{Model Size} & \textbf{\begin{tabular}[c]{@{}c@{}}ROI Velocity\\ ($N_{break}$)\end{tabular}} & \textbf{\begin{tabular}[c]{@{}c@{}}Green Efficiency\\ ($IPW$)\end{tabular}} & \textbf{\begin{tabular}[c]{@{}c@{}}System Density\\ ($\rho_{sys}$)\end{tabular}} & \textbf{\begin{tabular}[c]{@{}c@{}}Quantization Fidelity\\ ($Q_{ret}$)\end{tabular}} & \textbf{\begin{tabular}[c]{@{}c@{}}Cold-Start Tax\\ ($C_{tax}$)\end{tabular}} \\ \midrule
\textbf{LLaMa} & \textbf{1B} & \textbf{14 Reqs} & 0.45 & \textbf{6,930 Tok/s/GB} & \textbf{100.6\%} & \textbf{183x} \\
 & 3B & 33 Reqs & 0.27 & 1,336 Tok/s/GB & 99.8\% & 184x \\
 & 7B & 43 Reqs & 0.15 & 387 Tok/s/GB & 100.3\% & 230x \\ \midrule
\textbf{Qwen} & 1B & 21 Reqs & \textbf{0.48} & 6,942 Tok/s/GB & 99.6\% & 179x \\
 & 3B & 28 Reqs & 0.23 & 1,419 Tok/s/GB & 97.3\% & 188x \\
 & 7B & 39 Reqs & 0.14 & 394 Tok/s/GB & 99.5\% & 237x \\ \bottomrule
\end{tabular}%
}
\caption{Lifecycle efficiency frontier on legacy T4 hardware. Median INT4 results (20 runs, three tasks) across $N_{break}$, $IPW$, $\rho_{sys}$, $Q_{ret}$, and $C_{tax}$ show that compact ($<2$B) models consistently outperform larger tiers in ROI velocity, system density, and energy-normalized intelligence.}
\label{tab:hero_framework}
\end{table*}

\subsection{Dataset Analysis}
\label{Datasets}

To evaluate \textsc{Edge-Eval}, we use three representative datasets—\texttt{Summarization (XSum)} \cite{narayan-etal-2018-dont}, \texttt{RAG (SQuAD)} \cite{rajpurkar-etal-2016-squad}, and \texttt{Conversational Agent (UltraChat)} \cite{ding2023enhancing}. \texttt{XSum} contains $\sim$227K news articles (204K/11K/11K train/val/test) paired with single-sentence summaries—modeling long-context document compression under constrained inference budgets. \texttt{SQuAD} v1.1 provides $\sim$100K QA pairs (87K/10K train/val), adapted into a retrieval-grounded generation setup to simulate knowledge-intensive enterprise reasoning. \texttt{UltraChat} comprises $\sim$1.5M multi-turn dialogues—reflecting latency-sensitive conversational deployment. We follow an 70/15/15 train/validation/test split, limiting fine-tuning to 5K–10K training samples per task while reserving the full validation and test sets for evaluation.

\subsection{Evaluation Metrics}
\label{evaluation}

To evaluate \textsc{Edge-Eval}, we implement a three-pass evaluation loop and report mean, median, and standard deviation across task-specific metrics: for \texttt{RAG}, NLI Entailment (Context $\rightarrow$ Generation) \cite{honovich2022true} and ROUGE-L \cite{lin2004rouge}; for \texttt{Summarization}, NLI Non-Contradiction and ROUGE-L; and for \texttt{Conversational Agents}, LLM-as-a-Judge (GPT-4o) ratings \cite{zheng2023judging} on Helpfulness and Safety (1–10 Likert scale). Based on the lifecycle variables defined in Section~\ref{sec:methodology}, we formalize five deployment metrics. \emph{Economic Break-Even} computes the traffic volume required for local adaptation to undercut API costs, 
$N_{break} = \frac{C_{train} + C_{setup}}{C_{api} - C_{infer}}$, 
where $C_{train}$ denotes adaptation cost, $C_{setup}$ infrastructure overhead, $C_{api}$ per-request API cost, and $C_{infer}$ local inference cost per request. 
\emph{Intelligence-Per-Watt} measures task-normalized reasoning efficiency, 
$IPW = \frac{\mathcal{S}_{task} \cdot \alpha}{E_{req}}$, 
where $\mathcal{S}_{task}$ is normalized task performance, $\alpha$ a task-complexity scaling factor, and $E_{req}$ energy consumed per request. 
\emph{System Density} quantifies hardware utilization efficiency, 
$\rho_{sys} = \frac{\mathcal{T}_{put}}{M_{vram}}$, 
where $\mathcal{T}_{put}$ denotes sustained token throughput and $M_{vram}$ allocated GPU memory in GB. 
\emph{Cold-Start Tax} captures the relative energy overhead of model loading, 
$C_{tax} = \frac{E_{load}}{E_{infer}}$, 
with $E_{load}$ representing model loading energy and $E_{infer}$ steady-state inference energy. 
Finally, \emph{Quantization Fidelity} measures reasoning retention under 4-bit compression, 
$Q_{ret} = \left(\frac{\mathcal{S}_{INT4}}{\mathcal{S}_{FP16}}\right)\times100\%$, 
where $\mathcal{S}_{INT4}$ and $\mathcal{S}_{FP16}$ denote task scores under INT4 and FP16 precision, respectively. 

\begin{table}[t]
\centering
\scriptsize
\resizebox{\columnwidth}{!}{%
\begin{tabular}{@{}lllcc@{}}
\toprule
\textbf{Family} & \textbf{Size} & \textbf{Method} & \textbf{Median Energy (kWh)} & \textbf{Ratio} \\ \midrule
LLaMa 3.2 & 1B & LoRA-FP16 & 0.039 [0.025-0.045] & 1.0$\times$ \\
 &  & QLoRA-INT4 & 0.251 [0.231-0.355] & 6.4$\times$ \\ \midrule
LLaMa 3.2 & 3B & LoRA-FP16 & 0.171 [0.119-0.244] & 1.0$\times$ \\
 &  & QLoRA-INT4 & 0.511 [0.156-0.612] & 3.0$\times$ \\ \midrule
LLaMa 3.1 & 7B & LoRA-FP16 & 0.244 [0.235-0.251] & 1.0$\times$ \\
 &  & QLoRA-INT4 & 0.552 [0.463-0.691] & 2.3$\times$ \\ \midrule
Qwen 2.5 & 1.5B & LoRA-FP16 & 0.129 [0.096-0.185] & 1.0$\times$ \\
 &  & QLoRA-INT4 & 0.301 [0.295-0.433] & 2.3$\times$ \\ \bottomrule
Qwen 2.5 & 3B & LoRA-FP16 & 0.153 [0.120-0.177] & 1.0$\times$ \\
 &  & QLoRA-INT4 & 0.359 [0.326-0.408] & 2.3$\times$ \\ \bottomrule
Qwen 2.5 & 7B & LoRA-FP16 & 0.243 [0.082-0.386] & 1.0$\times$ \\
 &  & QLoRA-INT4 & 0.563 [0.492-0.633] & 2.3$\times$ \\ \bottomrule
\end{tabular}%
}
\caption{Adaptation energy asymmetry across precision regimes. Median training energy and carbon cost (20 runs, IQR shown) reveal that QLoRA—despite reducing VRAM—incurs up to $6.4\times$ higher energy for 1B models, exposing a divergence between memory and carbon efficiency on legacy hardware.}
\label{tab:energy_matrix}
\end{table}

\begin{figure*}[t]
\centering
\begin{subfigure}{0.32\textwidth}
    \centering
    \includegraphics[width=\linewidth]{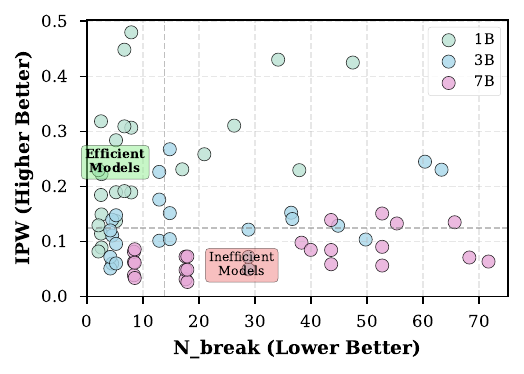}
    \caption{ROI–Efficiency Quadrant}
    \label{fig:roi_efficiency}
\end{subfigure}
\hfill
\begin{subfigure}{0.32\textwidth}
    \centering
    \includegraphics[width=\linewidth]{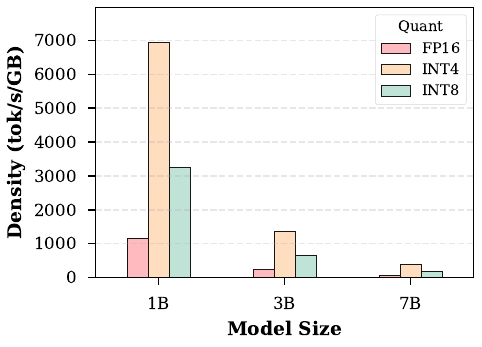}
    \caption{System Density}
    \label{fig:infrastructure_density}
\end{subfigure}
\hfill
\begin{subfigure}{0.32\textwidth}
    \centering
    \includegraphics[width=\linewidth]{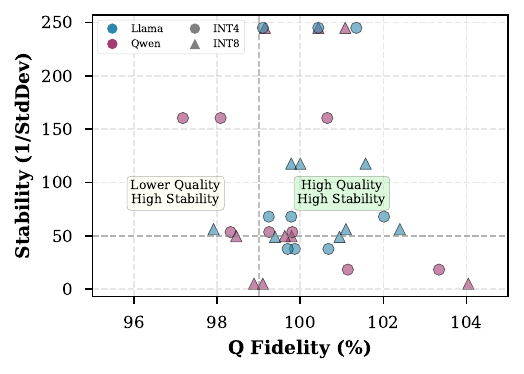}
    \caption{Quality–Stability Trade-off}
    \label{fig:quality_stability}
\end{subfigure}
\caption{Multidimensional efficiency under legacy deployment--compact ($<2$B) models form the ROI–IPW efficiency frontier, INT4 enables a strong throughput-per-GB hardware multiplier, and quantization fidelity remains stable with controlled variance.}
\label{fig:multidimensional_analysis}
\end{figure*}

\section{Result Analysis}

\subsection{Benchmark Analysis}

The results in Table~\ref{tab:hero_framework} reveal a clear efficiency frontier favoring the $<2$B \emph{parameter class} across economic, ecological, and infrastructure dimensions. The LLaMa-3.2-1B configuration achieves the fastest ROI velocity ($14$ requests), the highest system density ($6{,}930$ tokens/s/GB), and near-perfect quantization fidelity ($100.6\%$), while maintaining competitive green efficiency. Similarly, Qwen-1.5B attains the highest $IPW$ (0.48), confirming that compact models maximize reasoning-per-watt under constrained hardware. In contrast, 7B models exhibit 3–5$\times$ lower density and slower economic recovery despite marginal quality gains, indicating diminishing deployment returns at scale. Table~\ref{tab:energy_matrix} further exposes a critical lifecycle asymmetry--although QLoRA reduces memory footprint, it increases adaptation energy by up to $6.4\times$ for 1B models and $\sim2$–$3\times$ for larger tiers, demonstrating that memory efficiency does not guarantee carbon efficiency. Notably, post-training quantization incurs negligible overhead ($<0.2\%$ of training energy), reinforcing its practicality for inference optimization. 

\subsection{Multidimensional Efficiency Dynamics}

Figure~\ref{fig:multidimensional_analysis} synthesizes economic, infrastructure, and quality dimensions of deployment viability. The ROI–Efficiency quadrant (see Fig.~\ref{fig:roi_efficiency}) reveals a pronounced efficiency frontier, where LLaMa-3.2-1B occupies the top-left region—achieving rapid break-even (median: 14 requests) while sustaining up to $3\times$ higher intelligence-per-watt than 7B baselines—indicating that compact models maximize both capital recovery speed and energy-normalized reasoning. The infrastructure density analysis (see Fig.~\ref{fig:infrastructure_density}) further demonstrates a hardware multiplier effect: INT4-quantized 1B models exceed $6{,}900$ tokens/s/GB, representing up to $17\times$ higher service capacity relative to 7B variants, thereby transforming legacy Tesla T4 accelerators into high-throughput inference nodes. Finally, the quality–stability trade-off (see Fig.~\ref{fig:quality_stability}) shows that LLaMa models maintain near-perfect quantization fidelity (99\%–101\%) with controlled variance shifts, whereas larger and denser configurations exhibit increased output instability, particularly in conversational settings. 

\begin{table}[t]
\centering
\scriptsize
\scriptsize
\resizebox{\columnwidth}{!}{%
\begin{tabular}{@{}llccccc@{}}
\toprule
\textbf{Family} & \textbf{Size} & \textbf{Precision} & \textbf{Throughput} & \textbf{Speedup} & \textbf{Energy/req} & \textbf{Savings} \\ \midrule
LLaMa 3.2 & 1B & FP16 & 2,235 & 1.0x & 6.45 J & - \\
 & & \textbf{INT4} & \textbf{4,331} & \textbf{1.94x} & \textbf{2.50 J} & \textbf{61\%} \\ \midrule
LLaMa 3.2 & 3B & FP16 & 1,374 & 1.0x & 12.67 J & - \\
 & & \textbf{INT4} & \textbf{2,506} & \textbf{1.82x} & \textbf{5.39 J} & \textbf{57\%} \\ \midrule
Qwen 2.5 & 7B & FP16 & 948 & 1.0x & 20.68 J & - \\
 & & \textbf{INT4} & \textbf{1,723} & \textbf{1.82x} & \textbf{8.90 J} & \textbf{57\%} \\ \bottomrule
\end{tabular}%
}
\caption{Inference efficiency under INT4 on Tesla T4--1.8–1.9$\times$ throughput gains and 57\%–61\% energy reduction versus FP16 (median over 20 runs, three tasks), confirming low-bit inference as a hardware multiplier under constrained infrastructure.}
\label{tab:inference_matrix}
\end{table}

\begin{table}[t]
\centering
\scriptsize
\resizebox{\columnwidth}{!}{%
\begin{tabular}{@{}llcccc@{}}
\toprule
\textbf{Family} & \textbf{Task} & \textbf{FP16 Score} & \textbf{INT4 Score} & \textbf{Retention} & \textbf{STD $\Delta$} \\ \midrule
\textbf{LLaMa} & Chat & $7.31 \pm 0.04$ & $7.32 \pm 0.04$ & \textbf{100.1\%} & $-6.1\%$ \\
 & RAG & $0.75 \pm 0.01$ & $0.75 \pm 0.01$ & \textbf{100.4\%} & \textcolor{red}{$+65.9\%$} \\
 & Summ & $0.86 \pm 0.02$ & $0.85 \pm 0.02$ & 98.6\% & $-9.7\%$ \\ \midrule
\textbf{Qwen} & Chat & $7.42 \pm 0.10$ & $7.25 \pm 0.14$ & 97.7\% & $+45.6\%$ \\
 & RAG & $0.77 \pm 0.01$ & $0.76 \pm 0.01$ & 98.7\% & $+20.5\%$ \\
 & Summ & $0.84 \pm 0.01$ & $0.85 \pm 0.02$ & \textbf{100.2\%} & $+25.3\%$ \\ \bottomrule
\end{tabular}%
}
\caption{Quantization fidelity under INT4 across 400 evaluation runs per task (20$\times$20). Retention = (INT4/FP16)$\times$100\% and STD $\Delta$ captures variance shift, showing near-lossless retention for compact models with architecture-dependent stability sensitivity.}
\label{tab:quant_robustness}
\end{table}

\begin{table}[t]
\centering
\scriptsize
\resizebox{\columnwidth}{!}{%
\begin{tabular}{@{}lllcccc@{}}
\toprule
\textbf{Task} & \textbf{Metric} & \textbf{Family} & \textbf{Size} & \textbf{FP16} & \textbf{INT4} & \textbf{Retention} \\ \midrule
\multirow{6}{*}{Chat} & \multirow{6}{*}{\begin{tabular}[c]{@{}l@{}}Helpfulness\\(1-10)\end{tabular}} 
& & \textbf{1B} & $6.67 \pm 0.08$ & $6.72 \pm 0.05$ & \textcolor{green!60!black}{+0.8\%} \\
& & \textbf{LLaMa} & 3B & $7.58 \pm 0.01$ & $7.57 \pm 0.04$ & -0.2\% \\
& & & 7B & $7.68 \pm 0.02$ & $7.67 \pm 0.03$ & -0.1\% \\
\cline{3-7}
& & & 1B & $7.19 \pm 0.16$ & $7.34 \pm 0.17$ & +2.0\% \\
& & Qwen & 3B & $7.46 \pm 0.07$ & $6.75 \pm 0.23$ & \textcolor{red}{\textbf{-9.5\%}} \\
& & & 7B & $7.62 \pm 0.06$ & $7.66 \pm 0.02$ & +0.6\% \\ \midrule
\multirow{6}{*}{RAG} & \multirow{6}{*}{\begin{tabular}[c]{@{}l@{}}Entailment\\(0-1)\end{tabular}}
& & \textbf{1B} & $0.74 \pm 0.01$ & $0.76 \pm 0.02$ & +2.7\% \\
& & \textbf{LLaMa} & 3B & $0.77 \pm 0.01$ & $0.77 \pm 0.01$ & $\pm 0.0\%$ \\
& & & 7B & $0.74 \pm 0.00$ & $0.73 \pm 0.01$ & -1.4\% \\
\cline{3-7}
& & & 1B & $0.76 \pm 0.01$ & $0.75 \pm 0.01$ & -2.0\% \\
& & Qwen & 3B & $0.77 \pm 0.01$ & $0.77 \pm 0.01$ & +0.7\% \\
& & & 7B & $0.79 \pm 0.01$ & $0.77 \pm 0.01$ & -2.5\% \\ \midrule
\multirow{6}{*}{Summ} & \multirow{6}{*}{\begin{tabular}[c]{@{}l@{}}ROUGE-L\\(0-1)\end{tabular}}
& & \textbf{1B} & $0.77 \pm 0.01$ & $0.76 \pm 0.01$ & -0.7\% \\
& & \textbf{LLaMa} & 3B & $0.90 \pm 0.03$ & $0.86 \pm 0.02$ & -4.4\% \\
& & & 7B & $0.92 \pm 0.01$ & $0.93 \pm 0.02$ & +1.1\% \\
\cline{3-7}
& & & 1B & $0.79 \pm 0.01$ & $0.79 \pm 0.01$ & $\pm 0.0\%$ \\
& & Qwen & 3B & $0.86 \pm 0.01$ & $0.86 \pm 0.02$ & $\pm 0.0\%$ \\
& & & 7B & $0.89 \pm 0.01$ & $0.89 \pm 0.02$ & +0.6\% \\ \bottomrule
\end{tabular}%
}
\caption{Task-level quality under INT4 vs FP16 across Chat, RAG, and Summarization (400 runs per task). Median $\pm$ StdDev shows near-baseline retention for compact models, with deviations highlighting task-and architecture-specific sensitivity.}
\label{tab:quality_benchmark}
\end{table}

\begin{figure*}[t]
\centering

\begin{subfigure}{0.32\textwidth}
    \centering
    \includegraphics[width=\linewidth]{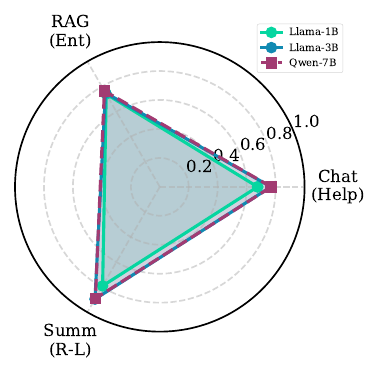}
    \caption{Multi-Task Profile}
\end{subfigure}
\hfill
\begin{subfigure}{0.32\textwidth}
    \centering
    \includegraphics[width=\linewidth]{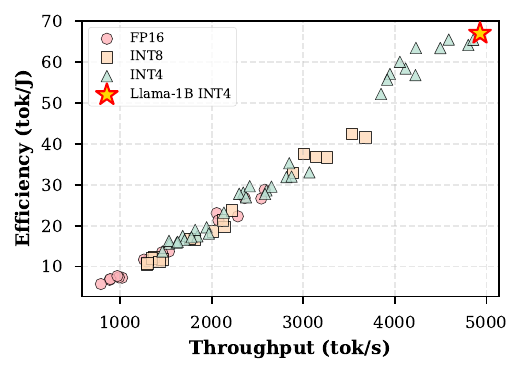}
    \caption{Pareto Frontier}
\end{subfigure}
\hfill
\begin{subfigure}{0.32\textwidth}
    \centering
    \includegraphics[width=\linewidth]{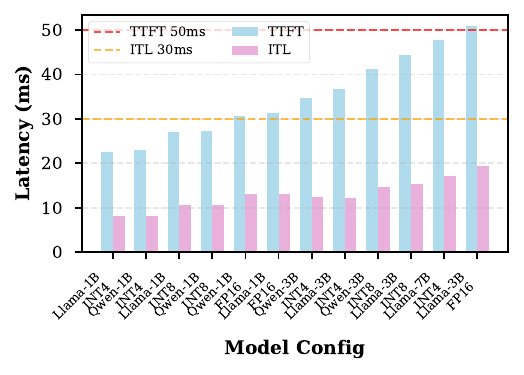}
    \caption{Latency Heatmap}
\end{subfigure}

\vspace{0.4cm}

\begin{subfigure}{0.32\textwidth}
    \centering
    \includegraphics[width=\linewidth]{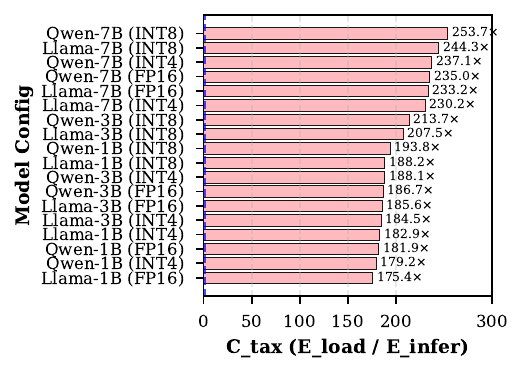}
    \caption{Cold-Start Tax}
\end{subfigure}
\hfill
\begin{subfigure}{0.32\textwidth}
    \centering
    \includegraphics[width=\linewidth]{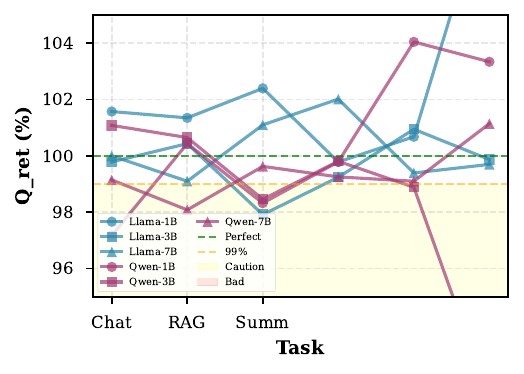}
    \caption{Quantization Fidelity}
\end{subfigure}
\hfill
\begin{subfigure}{0.32\textwidth}
    \centering
    \includegraphics[width=\linewidth]{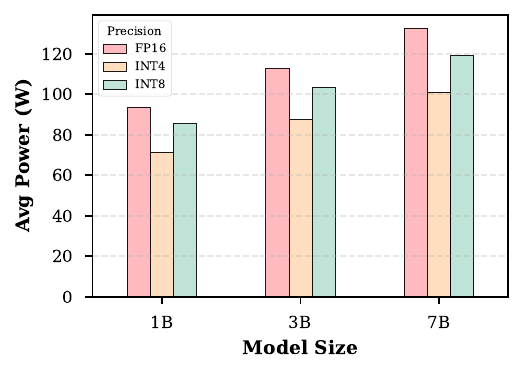}
    \caption{Power Stability}
\end{subfigure}
\caption{Systems-level deployment landscape on legacy T4 hardware. Compact ($<2$B) INT4 models consistently dominate the efficiency frontier—balancing multi-task performance, Pareto-optimal throughput–energy trade-offs, low latency, $>99\%$ quantization retention, and stable $\sim$35W power—while larger models remain lifecycle-dominated.}
\label{fig:systems_landscape}
\end{figure*}

\subsection{Inference and Quality Trade-offs}

Tables~\ref{tab:inference_matrix}--\ref{tab:quality_benchmark} jointly characterize the inference–quality balance under INT4 deployment. The Inference Performance Matrix (see Table~\ref{tab:inference_matrix}) demonstrates a consistent hardware multiplier effect--INT4 nearly doubles throughput (1.82-1.94$\times$ speedup) while reducing per-request energy by 57\%–61\%, confirming that low-bit inference materially improves energy-normalized service capacity on Tesla T4 hardware. Therefore, these efficiency gains do not systematically degrade task quality. As shown in Table~\ref{tab:quant_robustness}, LLaMa models retain 99\%–101\% performance across tasks with tightly controlled variance, whereas Qwen exhibits greater instability in conversational settings (+45.6\% standard deviation shift), indicating model-dependent quantization sensitivity. The comprehensive benchmark in Table~\ref{tab:quality_benchmark} further reveals that compact (1B-3B) models preserve or even slightly improve task scores under INT4, while occasional degradations (e.g., Qwen-3B Chat, 9.5\%) remain localized rather than systemic.

\subsection{Systems-Level Deployment Dynamics}

Figures~\ref{fig:systems_landscape} shows the systems-level behavior underlying the efficiency frontier. The multi-task radar profile confirms that LLaMa-3.2-1B (INT4) maintains balanced performance across Chat, RAG, and Summarization, forming a near-circular capability shape indicative of stable generalization at low cost. The Pareto frontier further demonstrates structural dominance--1B configurations occupy the optimal lower-right region (high throughput, low energy), while 7B models remain Pareto-inferior regardless of quantization strategy. Latency analysis reveals a non-linear deployment advantage—INT4 reduces inter-token latency disproportionately for compact models, enabling sub-10ms ITL and TTFT below 50ms, thereby satisfying real-time interaction thresholds. Operational constraints reinforce this asymmetry--cold-start taxes of 179–237$\times$ render scale-to-zero economically infeasible for low-traffic workloads, and although 4-bit quantization safely preserves reasoning fidelity (most configurations exceeding the 99\% threshold), larger models still incur higher steady-state power draw. The power profile highlights an additional edge advantage--LLaMa-3.2-1B sustains a stable $\sim$35W envelope under inference, supporting fanless deployment and improved reliability in constrained environments.

\section{Conclusion}

This work introduced \textsc{Edge-Eval}, a lifecycle-oriented benchmarking framework designed to close the \emph{Deployment--Evaluation Gap} in industry LLM assessment. Through LLaMA and Qwen variants across adaptation, quantization, and inference on legacy Tesla T4 hardware, we demonstrated that compact ($<2$B) models consistently dominate larger baselines in ROI velocity, energy-normalized intelligence, system density, and latency stability under INT4 deployment. Our findings reveal two key insights--(i) small models form a clear efficiency frontier under constrained infrastructure, and (ii) memory-efficient training (e.g., QLoRA) does not necessarily imply energy or carbon efficiency. 

\section*{Limitations}

Our work focuses on legacy NVIDIA Tesla T4 accelerators and low-batch deployment settings; results may differ on modern architectures (e.g., Hopper-class GPUs) or high-throughput cloud serving environments. We evaluate two model families and three industry-representative tasks, which, while diverse, do not exhaust the space of domain-specific workflows. Energy measurements rely on GPU-level power telemetry and may not capture full system-level overheads (e.g., CPU, networking). At last, economic assumptions (e.g., API pricing and carbon intensity factors) reflect current estimates and may evolve over time, affecting absolute break-even thresholds.

\section*{Ethics Statement}

This work evaluates operational efficiency rather than proposing new model capabilities. All experiments are conducted on publicly available open-weight models and widely used benchmark datasets. Through emphasizing energy consumption, carbon footprint, and infrastructure efficiency, \textsc{Edge-Eval} aligns with responsible and sustainable AI principles. However, improved deployment efficiency may lower the barrier to large-scale model use; practitioners must ensure compliance with data governance, privacy, and responsible deployment standards in real-world industry applications.

\bibliography{custom}

\end{document}